\documentclass[letterpaper]{article} % DO NOT CHANGE THIS
\usepackage[preprint]{aaai2027}
\usepackage[hyphens]{url}  % DO NOT CHANGE THIS
\usepackage{graphicx} % DO NOT CHANGE THIS
\usepackage{natbib}  % DO NOT CHANGE THIS AND DO NOT ADD ANY OPTIONS TO IT
\usepackage{caption} % DO NOT CHANGE THIS AND DO NOT ADD ANY OPTIONS TO IT
\usepackage{algorithm}
\usepackage{algorithmic}
\usepackage{multirow}
\makeatletter

\newcommand{\Rmnum}[1]{\expandafter\@slowromancap\romannumeral #1@}
\makeatother
\usepackage{amssymb}
\usepackage{newfloat}
\usepackage{listings}
\DeclareCaptionStyle{ruled}{labelfont=normalfont,labelsep=colon,strut=off} % DO NOT CHANGE THIS
\floatstyle{ruled}
\newfloat{listing}{tb}{lst}{}
\floatname{listing}{Listing}

\usepackage{booktabs}

\title{IntactWorld: Joint World Modeling with Intact Features}
\author{
Boming Tan$^{1}$, Xiangdong Zhang$^{2}$, Yan Xia$^{1}$, Qi Zhu$^{3}$, Deyi Ji$^{3}$, Xue Yang$^{2}$, Shaofeng Zhang$^{1}$\thanks{Shaofeng Zhang is the corresponding author.} \\
}
\affiliations{
\textsuperscript{\rm 1}University of Science and Technology of China, 
\textsuperscript{\rm 2}Shanghai Jiao Tong University, 
\textsuperscript{\rm 3}KOKONI 3D, Moxin Technology\\
    bomingtan@foxmail.com, sfzhang@ustc.edu.cn
}

\begin{document}

\maketitle

\begin{abstract}
\label{abs}
While recent video generation models synthesize highly realistic visuals, they lack a genuine understanding of intrinsic real-world logic. Existing methods attempt to understand the world by internalizing diverse world knowledge, yet constrained by computational overhead or dimensionality alignment, their learning processes inevitably compress features, causing a severe loss of structural information. To address this, we propose \textbf{IntactWorld}, a \textbf{Joint World Modeling Architecture} utilizing uncompressed \textbf{Intact Features}. Since data naturally reside on a low-dimensional manifold within a high-dimensional space, predicting the flow velocity $v$ within this uncompressed high-dimensional space induces a severe manifold gap. To successfully eliminate this optimization bottleneck, our framework instead predicts the clean feature $x_0$ at intermediate layers. Furthermore, to mitigate the computational overhead of incorporating complete world knowledge, we introduce a \textit{Full-to-Compact Training Paradigm}. By replacing raw full features with highly refined CLS tokens, this paradigm enables efficient single-branch guidance, reducing spatial memory consumption by 11.4\% and cutting inference latency by 43.8\%. Extensive evaluations demonstrate the effectiveness of IntactWorld, outperforming established baselines by 2.46 points on the VBench 2.0 benchmark.
\end{abstract}

\section{Introduction}
\label{intro}

Recent advancements in Latent Diffusion Models (LDMs) \cite{rombach2022highresolutionimagesynthesislatent} and Diffusion Transformers (DiT) \cite{peebles2023scalablediffusionmodelstransformers} have significantly propelled video generation. Foundation models such as CogVideoX \cite{yang2025cogvideoxtexttovideodiffusionmodels}, HunyuanVideo \cite{kong2025hunyuanvideosystematicframeworklarge}, and Wan2.1 \cite{wan2025wanopenadvancedlargescale} can now synthesize highly realistic visual sequences. Despite this impressive visual fidelity, these architectures usually operate as pixel-level pattern matchers. Optimized primarily for surface-level statistical distributions, they often struggle to capture the intrinsic logic of the real world. This fundamental limitation is reflected by their performance on world-centric benchmarks such as VBench 2.0 \cite{zheng2025vbench20advancingvideogeneration}. These limitations suggest that generative frameworks should move beyond pixel-level dependencies and incorporate richer world knowledge.

Existing methods \cite{lin2025uniworld,videoworldsimulators2024,yuan2026inferencetimephysicsalignmentvideo,yang2026omnivideo2scalingmllmconditioned,zhu2024exploringpretrainedtexttovideodiffusion} have sought to equip video generators with external priors to facilitate genuine world modeling. Approaches like DreamWorld \cite{tan2026dreamworldunifiedworldmodeling} extend projection layers to explicitly internalize multi-modal knowledge. To manage the resulting computational overhead, they heavily rely on dimensionality reduction techniques such as Principal Component Analysis (PCA) \cite{kouzelis2026boostinggenerativeimagemodeling}. However, this compression erases structural details, specifically spatial and motion information, causing the model to overfit to generalized semantics. Alternatively, discarding PCA to train directly on uncompressed full features introduces a critical optimization bottleneck. Diffusion models typically predict the velocity vector $v$. According to the manifold hypothesis \cite{li2026basicsletdenoisinggenerative}, data naturally resides on a low-dimensional manifold within a high-dimensional space. Therefore, directly predicting $v$ in this uncompressed high-dimensional space induces significant optimization loss, which degrades the overall generative capacity of the model (as detailed in Section 4.4).

To address this gap, we propose \textbf{IntactWorld}, a \textbf{Joint World Modeling Architecture} utilizing \textbf{Intact Features}. Inspired by DreamWorld, we extend projection layers to internalize knowledge, but incorporate uncompressed full DINOv2 \cite{oquab2023dinov2} and VGGT \cite{wang2025vggtvisualgeometrygrounded} features. To circumvent the manifold gap bottleneck caused by directly predicting $v$ in a high-dimensional space~\cite{li2026basicsletdenoisinggenerative}, our model avoids directly predicting velocity for the world features. Instead, it predicts clean world features $x_0$ while retaining velocity prediction for the VAE latents, and computes the training loss in velocity space. Furthermore, since intermediate network layers preserve richer structural information, we predict $x_0$ from these intermediate layers rather than the final layer.

However, incorporating external world knowledge inevitably increases inference costs. For instance, DreamWorld employs a Multi-Source Inner-Guidance mechanism, which requires constructing multiple auxiliary branches on top of standard classifier-free guidance(CFG)~\cite{vincent2011connection,ho2022classifierfreediffusionguidance}. To address this, we propose a \textit{Full-to-Compact Training Paradigm}. Building upon the full-feature learning of Stage \Rmnum{1}, Stage \Rmnum{2} uses compact CLS tokens and concatenates them with the VAE latents in the hidden layers. Since these CLS tokens have already internalized the world knowledge and are highly condensed, they inherently provide strong guidance. Consequently, instead of requiring multiple branches, we only need to construct a single additional CLS feature branch alongside standard CFG, which drastically reduces inference overhead. \textbf{Our main contributions are as follows:}

\textbf{1)} We propose \textbf{IntactWorld}, a \textbf{Joint World Modeling Architecture} utilizing \textbf{Intact Features}. By predicting the clean features $x_0$ and computing the loss in the velocity space, it alleviates the bottleneck of the manifold gap and successfully internalizes complete world features.

\textbf{2)} We drastically reduce inference costs through a novel \textbf{Full-to-Compact Training Paradigm} and \textit{Compact Inner-Guidance} mechanism. By abstracting world knowledge into compact CLS tokens, we achieve highly efficient generation with only a single additional branch atop standard CFG.

\textbf{3)} Extensive experiments demonstrate that IntactWorld significantly outperforms baselines and DreamWorld, establishing new records on the VBench and VBench 2.0.

\section{Related Works}
\paragraph{Video Diffusion Models} The development of video generation builds upon latent diffusion models (LDMs)~\cite{rombach2022highresolutionimagesynthesislatent,zhang2025show1marryingpixellatent,ma2025lattelatentdiffusiontransformer}, which operate in a compressed latent space to enable efficient synthesis while preserving perceptual quality. Early approaches such as VideoCrafter1~\cite{chen2023videocrafter1opendiffusionmodels} and ModelScope~\cite{wang2023modelscopetexttovideotechnicalreport} extend UNet-based architectures with interleaved spatial and temporal attention modules to capture frame dependencies. Subsequently, the field has shifted toward Diffusion Transformers (DiT)~\cite{peebles2023scalablediffusionmodelstransformers}, replacing convolutional U-Nets with pure transformer backbones that better scale with model size and sequence length. Foundational models including Sora \cite{videoworldsimulators2024}, CogVideoX~\cite{yang2025cogvideoxtexttovideodiffusionmodels} , HunyuanVideo~\cite{kong2025hunyuanvideosystematicframeworklarge} and Wan2.1 \cite{wan2025wanopenadvancedlargescale} leverage large-scale DiT architectures, often combined with Flow Matching~\cite{lipman2023flowmatchinggenerativemodeling} frameworks, which enables straighter and faster generation while maintaining high fidelity.

\paragraph{World Modeling}
World models~\cite{https://doi.org/10.5281/zenodo.1207631} are already being developed in the field of video generation. The JEPA family~\cite{bardes2024revisitingfeaturepredictionlearning,huang2026vjepavariationaljointembedding,garrido2024learningleveragingworldmodels,assran2025vjepa2selfsupervisedvideo,bardes2024revisitingfeaturepredictionlearning,maes2026leworldmodelstableendtoendjointembedding} shifts from pixel reconstruction to prediction in a learned latent space, enabling zero-shot robotic planning at scale. Meanwhile, action-conditioned generative models such as Genie~\cite{bruce2024geniegenerativeinteractiveenvironments}, Genie2~\cite{parkerholder2024genie2} and the VideoWorld series~\cite{ren2025videoworldexploringknowledgelearning,ren2026videoworld2} learn transferable knowledge from unlabeled videos by decoupling action dynamics from visual appearance. Moving beyond single-source knowledge, DreamWorld~\cite{tan2026dreamworldunifiedworldmodeling} proposes a unified joint world modeling paradigm that simultaneously integrates multiple world knowledge. To further address compounding errors in long-horizon generation, retrieval-augmented strategies~\cite{chen2026learningworldmodelsinteractive,wang2026mechanisticviewvideogeneration,wu2025geometryforcingmarryingvideo,xiao2026worldmemlongtermconsistentworld,lin2025uniworld} have been proposed, alongside a unified mechanistic view of state and dynamics. These world modeling principles have also been successfully extended to autonomous driving~\cite{li2025enhancingendtoendautonomousdriving,min2024driveworld4dpretrainedscene}, demonstrating broad applicability beyond simulated environments.
\paragraph{Manifold Gap}
The manifold hypothesis posits that natural data reside on a low-dimensional manifold embedded in a high-dimensional space, whereas noise or noised quantities lie off this manifold. Drawing on this perspective, JiT~\cite{li2026basicsletdenoisinggenerative} proposes directly predicting clean data ($x$-prediction), demonstrating that even networks with limited capacity can handle high-dimensional patches when the prediction respects manifold structure. Enforcing bottleneck or sparse structures is a common strategy for learning such low-dimensional representations~\cite{tishby2000informationbottleneckmethod,alemi2016deep,rifai2011contractive,makhzani2014ksparseautoencoders}. This principle has since been extended to address geometric interference, manifold-guided distillation, concept manifolds in sparse autoencoders, and manifold-aware video generation~\cite{kumar2026learningmanifoldunlockingstandard,farghly2025diffusionmodelsmanifoldhypothesis,roy2026manifoldgdtrainingfreehierarchicalmanifold,bhalla2026sparseautoencoderscaptureconcept,zheng2026manifoldawareexplorationreinforcementlearning,weng2026bridgingrestorationgenerationmanifolds}. Together, these works highlight the importance of respecting data geometry in generative modeling.
\begin{figure*}[tb!]
    \centering
    \includegraphics[width=\textwidth]{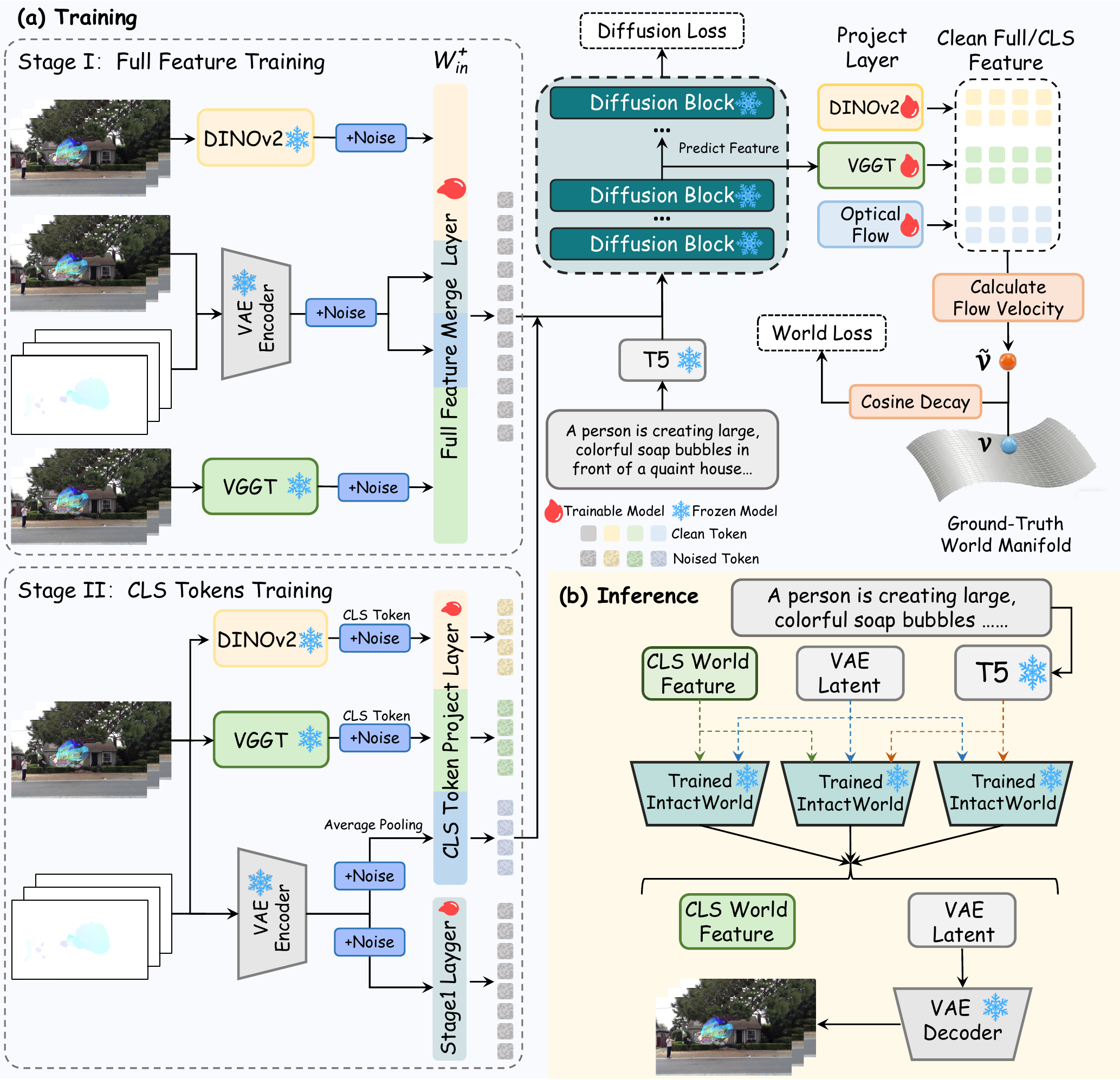}
    % \vspace{-20pt}
    \caption{\textbf{Overview of the IntactWorld Framework.} \textbf{(a) Training:} We employ a Full-to-Compact Training Paradigm. \textbf{Stage \Rmnum{1}} utilizes lossless full features, predicting the clean target $x_0$ to compute the World Loss in the $v$-space. \textbf{Stage \Rmnum{2}} trains exclusively on compact CLS tokens for semantic abstraction. \textbf{(b) Inference:} These condensed tokens enable Compact Inner-Guidance, achieving efficient generation via a single additional branch atop standard CFG.}
    \label{fig:overview}
    % \vspace{-5pt}
\end{figure*}
\section{Method}
\label{sec:method}
In this section, we present \textbf{IntactWorld}, a Joint World Modeling Architecture with intact features, which is illustrated in Figure \ref{fig:overview}. Sections 3.1 and 3.2 introduce Flow Matching preliminaries and feature preprocessing. Section 3.3 details the Stage \Rmnum{1} joint modeling with intact features to overcome the manifold gap. Section 3.4 outlines the Stage \Rmnum{2} feature abstraction into compact CLS tokens using a position-isolated RoPE strategy. Finally, Section 3.5 presents the Compact Inner-Guidance mechanism for highly efficient inference.
\subsection{Preliminaries}
\label{subsec:preli}
\paragraph{Flow Matching.} Flow Matching \cite{lipman2023flowmatchinggenerativemodeling} has emerged as a highly efficient continuous-time generative paradigm, powering recent state-of-the-art video models such as Wan2.1 \cite{wan2025wanopenadvancedlargescale}. Let $z_0 \sim p_{data}$ denote the real video latent and $z_1 \sim \mathcal{N}(0, I)$ be the standard Gaussian prior. Flow Matching constructs a continuous probability path $z_t = t z_1 + (1 - t) z_0$ for $t \in [0, 1]$, and optimizes a neural network $v_\theta$ to predict the constant flow velocity $z_1 - z_0$. The standard objective is defined as:
\begin{equation}
\mathcal{L}_{FM} = \mathbb{E}_{t, z_0, z_1, c} \left[ \| v_\theta(z_t, t, c) - (z_1 - z_0) \|^2 \right]
\end{equation}
where $c$ represents the textual condition. During inference, the video latent is recovered by numerically solving the ODE from $t=1$ to $t=0$.

\paragraph{Data Manifold and Prediction Targets.} While standard diffusion models typically predict noise or velocity, these targets struggle in high-dimensional dense feature spaces\cite{li2026basicsletdenoisinggenerative}. Under the manifold assumption, structured data $x_0$ resides on a low-dimensional manifold $\mathcal{M} \subset \mathbb{R}^D$ ($d \ll D$). In contrast, Gaussian noise $\epsilon$ and flow velocity $v = x_0 - \epsilon$ span the entire high-dimensional space $\mathbb{R}^D$. Consequently, predicting clean data ($x_0$) is highly tractable since it only requires retaining low-dimensional structural information. Conversely, predicting off-manifold quantities ($v$) forces limited-capacity networks to model the full high-dimensional distribution, inevitably leading to severe optimization conflicts when injecting dense external features.

\subsection{Preprocessing}
\label{subsec:prepro}
\paragraph{Motion Representation Transformation.} To explicitly capture temporal dynamics, we extract Optical Flow~\cite{teed2020raftrecurrentallpairsfield} fields $(u, v)$ and transform them into a 3-channel RGB format for standard video encoders. The motion magnitude $m$ and direction $\alpha$ are formulated as:
\begin{equation}
m = \min\left(1, \frac{\sqrt{u^2 + v^2}}{\gamma \sqrt{H^2 + W^2}}\right), \quad \alpha = \arctan2(v, u)
\end{equation}
where $\gamma$ scales magnitude sensitivity. These components modulate pixel-wise intensity and hue, respectively. A frozen 3D VAE then encodes these frames into a temporal latent $z_{temporal}$, effectively projecting the motion prior into the visual latent space.

\paragraph{Feature Distribution Alignment.} A distributional discrepancy exists between the generative VAE latent space and dense expert features from models like VGGT~\cite{wang2025vggtvisualgeometrygrounded} and DINOv2~\cite{oquab2023dinov2}. To mitigate this, we first match their resolutions to the video latents via spatial interpolation and temporal pooling. Next, we apply channel-wise standardization (zero mean, unit variance) to the full features. This aligns the heterogeneous representations onto a unified statistical manifold, critically mitigating the manifold gap and ensuring optimization stability during joint modeling. Without normalization, features from different experts can dominate the joint representation unevenly, which destabilizes training and weakens effective cross-modal fusion.

\subsection{Stage \Rmnum{1}: Full Feature Training}
\label{subsec:stage1}
\paragraph{Joint Modeling with Full Features.} Building a robust world model requires internalizing semantic, spatial, and temporal dynamics. While previous alignment paradigms compress multi-source features to reduce compute, this dimensional reduction inevitably erases precise spatial geometry and nuanced motion details. To preserve these world priors, we bypass compression and utilize the full dense features from DINOv2 and VGGT. Let $Z_{world} = [z_{sem}, z_{spa}, z_{temp}]$ denote the concatenated uncompressed world representations. The clean joint state is defined as $Z_0 = [z_{vae}, Z_{world}]$. Following the unified modeling paradigm \cite{tan2026dreamworldunifiedworldmodeling}, we integrate this composite tensor by expanding the input projection layer $W_{in}$ into $W_{in}^+$, where the extended weights are zero-initialized:
\begin{equation}
F_{in} = Z_t \cdot W_{in}^+ = [\tilde{z}_{vae}, \tilde{Z}_{world}] \cdot [W_{in}, \mathbf{0}]
\end{equation}
where $Z_t$ denotes the noised joint state at time step $t$. In practice, this design provides a smooth adaptation path from the pre-trained video generator to the joint world-modeling regime. It prevents the newly introduced feature channels from abruptly perturbing the original latent distribution at the beginning of training.
\paragraph{Manifold Gap.} Integrating full expert features drastically inflates the dimensionality of $Z_t$, introducing a profound optimization dilemma known as the manifold gap \cite{li2026basicsletdenoisinggenerative}. Standard Flow Matching models typically predict flow velocity $v$. However, $v$ is an off-manifold quantity that spans the high-dimensional space. Regressing these targets causes optimization conflicts and structural collapse. Conversely, the clean world feature $Z_{world}$ intrinsically resides on a low-dimensional manifold. Driven by this insight, we introduce a decoupled prediction mechanism. We retain $v$-prediction for VAE latents but reparameterize the network to directly predict the on-manifold data for the world priors:
\begin{equation}
[\hat{v}_{vae}, \hat{Z}_{world}] = net_\theta([\tilde{z}_{vae}, \tilde{Z}_{world}], t, c)
\end{equation}
By shifting the target to the structured world reality, the network acts as a bottleneck that suppresses high-dimensional noise and concentrates model capacity on recovering the fine-grained structure of world knowledge.

\paragraph{Optimization in Velocity Space.} While predicting $\hat{Z}_{world}$ facilitates optimization, computing the regression loss directly in the data space severs the connection to the ordinary differential equation (ODE) framework. To preserve this connection, we project the prediction back into the velocity space. For the world features, the ground-truth velocity $v_{world}$ and implied predicted velocity $\hat{v}_{world}$ are derived as:
\begin{equation}
v_{world} = \frac{\tilde{Z}_{world} - Z_{world}}{t}, \quad \hat{v}_{world} = \frac{\tilde{Z}_{world} - \hat{Z}_{world}}{t}
\end{equation}
Additionally, to prevent the strong structural constraints from overwhelming the pixel-level visual generation, we introduce a Cosine Decay mechanism $\lambda$.  The joint objective is defined as:
\begin{equation}
\mathcal{L}_{total} = \mathbb{E} \left[ \|\hat{v}_{vae} - v_{vae}\|^2 + \lambda \cdot \frac{1}{t^2} \|\hat{Z}_{world} - Z_{world}\|^2 \right]
\end{equation}
This formulation naturally induces a $1/t^2$ reweighting, which penalizes structural errors more heavily when noise is low as $t \to 0$, forcing the model to refine local details. Meanwhile, the Cosine Decay factor $\lambda$ gradually reduces this world guidance as training proceeds, ensuring the model internalizes world knowledge early while prioritizing high-fidelity visual generation in later stages.

\begin{table*}[tb!]
    \centering
    \setlength{\tabcolsep}{1.1pt} 
    \renewcommand{\arraystretch}{1.3}
    \begin{tabular}{l ccc ccc c cc c}
            \toprule
            \multirow{2}{*}[-2ex]{\textbf{\large Method}} & 
            \multicolumn{3}{c}{\textbf{ Temporal}} & 
            \multicolumn{3}{c}{\textbf{ Semantic}} & 
            \multicolumn{1}{c}{\textbf{ Spatial}} & 
            \multicolumn{2}{c}{\textbf{ Summary}} &
            \multirow{2}{*}[-2ex]{\textbf{ \shortstack{Overall\\Score}}} \\

            \cmidrule(lr){2-4} \cmidrule(lr){5-7} \cmidrule(lr){8-8} \cmidrule(lr){9-10}
            
            & \textbf{\shortstack{Subject\\Consistency}} 
            & \textbf{\shortstack{Background\\Consistency}} 
            & \textbf{\shortstack{Dynamic\\Degree}} 
            & \textbf{\shortstack{Object\\Class}} 
            & \textbf{\shortstack{Human\\Action}} 
            & \raisebox{1.2ex}{\textbf{Scene}} 
            & \textbf{\shortstack{Spatial\\Relationship}}
            & \textbf{\shortstack{Quality\\Score}}
            & \textbf{\shortstack{Semantic\\Score}}
            & \\ 
            \midrule 
            Wan2.1-T2V-1.3B & 91.83 & 94.71 & 65.00 & 76.09 & 74.60 & 20.03 & 62.37 & 79.81 & 65.43 & 76.93 \\
            Baseline & 93.59 & \underline{95.81} & 54.08 & 79.90 & 78.98 & 28.55 & 63.31 & 81.26 & 68.47 & 78.71 \\
            DreamWorld & \textbf{93.62} & 94.95 & \textbf{79.16} & \underline{81.32} & \underline{81.20} & \underline{29.71} & \textbf{70.47} & \underline{83.49} & \underline{70.89} & \underline{80.97} \\
            \midrule
            \textbf{\large IntactWorld (Ours)} & \underline{93.60} & \textbf{97.44} & \underline{78.94} & \textbf{82.63} & \textbf{83.40} & \textbf{31.34} & \underline{69.78} & \textbf{83.76} & \textbf{72.19} & \textbf{81.45} \\
            \bottomrule
        \end{tabular}%
    \caption{\textbf{Quantitative comparison on VBench.}  \textbf{Bold} and \underline{underline} indicate the best and second best results, respectively. IntactWorld achieves the best performance, clearly outperforming existing methods.}
    \label{tab:vbench}
\end{table*}
\begin{table*}[tb!]
    \centering
    \footnotesize 
    \renewcommand{\arraystretch}{1} 
    
    \begin{minipage}[t]{0.49\linewidth}
        \centering
        \setlength{\tabcolsep}{1.5pt}
        \begin{tabular}{lccccc}
            \toprule
            \multirow{2}{*}[-2pt]{\textbf{Method}} & 
            \multicolumn{4}{c}{\textbf{VBench 2.0 Dimensions}} & 
            \multirow{2}{*}[-0.5ex]{\textbf{Overall}} \\
            \cmidrule(lr){2-5}
            & \textbf{Com.} 
            & \textbf{Control.} 
            & \textbf{Hum. Fid.} 
            & \textbf{Physics} 
            & \\ 
            \midrule
            Wan2.1-T2V-1.3B        & 59.17 & 16.81 & 76.09 & \underline{55.85} & 50.77 \\
            Baseline   & \underline{62.80} & \textbf{18.41} & 77.09 & 54.51 & 51.18 \\
            DreamWorld          & 61.82 & 16.95 & \textbf{80.11} & 55.07 & \underline{52.97} \\
            \midrule
            \textbf{IntactWorld (Ours)}  & \textbf{64.83} & \underline{18.38} & \underline{77.72} & \textbf{57.91} & \textbf{53.64} \\
            \bottomrule
        \end{tabular}
        \caption{\textbf{VBench-2.0 quantitative results.} \textbf{Com.}: Commonsense; \textbf{Control.}: Controllability; \textbf{Hum. Fid.}: Human Fidelity. \textbf{Bold} and \underline{underline} denote the best and \underline{second-best} results.}
        \label{tab:vbench2}
    \end{minipage}\hfill
    \begin{minipage}[t]{0.49\linewidth}
        \centering
        \setlength{\tabcolsep}{1.0pt}
        \begin{tabular}{lcccccccc}
            \toprule
            \multirow{2}{*}[-2pt]{\textbf{Method}} & 
            \multicolumn{2}{c}{\textbf{Solid-Solid}} & 
            \multicolumn{2}{c}{\textbf{Solid-Fluid}} & 
            \multicolumn{2}{c}{\textbf{Fluid-Fluid}} & 
            \multicolumn{2}{c}{\textbf{Overall}} \\
            \cmidrule(lr){2-3} \cmidrule(lr){4-5} \cmidrule(lr){6-7} \cmidrule(lr){8-9}
            & \textbf{SA} & \textbf{PC} & \textbf{SA} & \textbf{PC} & \textbf{SA} & \textbf{PC} & \textbf{SA} & \textbf{PC} \\
            \midrule
            Wan2.1-T2V-1.3B       & 51.1 & 22.3 & 45.2 & 19.1 & 45.4 & 23.6 & 47.7 & 21.2 \\
            Baseline  & 46.2 & 18.2 & 47.8 & 22.6 & 50.9 & 23.6 & 45.1 & 20.9 \\
            DreamWorld            & 54.5 & 24.5 & 48.6 & 25.4 & \textbf{60.1} & 32.6 & 52.9 & 26.2 \\
            \midrule
            \textbf{IntactWorld (Ours)} & \textbf{55.9} & \textbf{25.8} & \textbf{53.4} & \textbf{27.3} & 58.1 & \textbf{32.7} & \textbf{55.2} & \textbf{27.6} \\
            \bottomrule
        \end{tabular}
        \caption{\textbf{Quantitative comparison on VideoPhy.} We report Semantic Adherence (SA) and Physical Commonsense (PC) scores. The best results are highlighted in \textbf{bold}.}
        \label{tab:videophy}
    \end{minipage}
\end{table*}

\subsection{Stage \Rmnum{2}: CLS Tokens Training}
\label{subsec:stage2}
\paragraph{Compact Internalization.} Existing world-guided generation frameworks typically require multiple expert guidance branches during inference~\cite{tan2026dreamworldunifiedworldmodeling}, resulting in substantial computational overhead. To address this issue, we extend classifier-free guidance (CFG)~\cite{vincent2011connection,dhariwal2021diffusionmodelsbeatgans,ho2022classifierfreediffusionguidance} with a single additional CLS branch. Since the compact CLS tokens $Z_{cls}$ already internalize the multi-source structural knowledge learned in Stage \Rmnum{1} in a highly condensed form, the model no longer requires separate semantic, spatial, and motion guidance branches during inference. This significantly reduces the number of forward passes while retaining effective world-guided generation capability.

\paragraph{Token Construction.} In this stage, we transition the network from full feature alignment to compact knowledge integration. The dense spatial and semantic features from VGGT~\cite{wang2025vggtvisualgeometrygrounded} and DINOv2~\cite{oquab2023dinov2} are explicitly replaced by their CLS tokens, $z_{spa}^{cls}$ and $z_{sem}^{cls}$. Specifically, both models extract $T$ CLS tokens, where $T$ corresponds to the temporal dimension of the latents. For the temporal prior, we apply spatial average pooling to the dense optical flow latents to obtain $T$ corresponding pseudo-CLS tokens. For the $i$-th temporal feature, this token is computed as:
$$z_{temp}^{cls, (i)} = \frac{1}{H \times W} \sum_{x, y} Z_{temp}^{(i, x, y)}$$
where $i \in \{1, 2, \dots, T\}$, and $H, W$ represent the spatial height and width of the latent representations.Let the concatenated world tokens be $Z'_{cls} = [z_{sem}^{cls}, z_{spa}^{cls}, z_{temp}^{cls}]$. To map these tokens into the transformer's hidden dimension $C$, we introduce a new projection layer $W_{cls}$. To ensure a stable transition, the projection weights for the VAE latents are strictly inherited from Stage \Rmnum{1}, whereas $W_{cls}$ is rigorously zero-initialized:
\begin{equation}
Z_{cls} = Z'_{cls} \cdot W_{cls}
\end{equation}
This zero-initialization ensures that the model progressively learns the compact representations without abruptly disrupting the visual backbone established in Stage \Rmnum{1}.

\paragraph{Sequence Concatenate and Positional Isolation.} After projection, the VAE hidden states $H_{vae} \in \mathbb{R}^{L_{vae} \times C}$ and the world tokens $Z_{cls} \in \mathbb{R}^{N_{cls} \times C}$ are concatenated along the sequence dimension to form the joint hidden state $H_{joint} \in \mathbb{R}^{(L_{vae} + N_{cls}) \times C}$. However, applying standard Rotary Positional Embeddings (RoPE) directly to this joint sequence would erroneously assign artificial spatial-temporal coordinates to the global $Z_{cls}$ tokens. To prevent this positional interference, we design a novel position-isolated RoPE strategy. Let $\Theta_{vae}$ denote the original rotary embedding tensor for the visual latents.~\cite{wu2025representationentanglementgenerationtraining}. We construct a position-neutral embedding for the CLS tokens using a tensor of ones, $\mathbf{1}_{cls}$, matching the sequence length $N_{cls}$. The new full rotary embedding is formulated by concatenating them along the sequence dimension:
\begin{equation}
\Theta_{full} = [\Theta_{vae}, \mathbf{1}_{cls}]
\end{equation}
This modification disables the rotational coordinate shifts for the CLS tokens. It ensures that the self-attention mechanism processes the compact world knowledge purely based on its inherent content, entirely free from spatial-temporal coordinate constraints.

\subsection{Compact Inner-Guidance}
\label{subsec:infer}
Standard classifier-free guidance~\cite{vincent2011connection,dhariwal2021diffusionmodelsbeatgans,ho2022classifierfreediffusionguidance} requires two forward branches. DreamWorld~\cite{tan2026dreamworldunifiedworldmodeling} extends this to a five-branch Multi-Source Inner-Guidance, which incurs prohibitive inference costs.

To address this, we introduce \textbf{Compact Inner-Guidance}. Since the unified tokens $Z_{cls}$ successfully internalize multi-source structural priors during Stage \Rmnum{2} training, they can act as a holistic world prior. By eliminating redundant modal branches, we achieve robust self-guidance using only three branches:
\begin{equation}
\hat{v}_{ours} = v_{joint} + w_{txt} \left( v_{joint} - v_{\neg txt} \right) + w_{cls} \left( v_{joint} - v_{\neg cls} \right)
\end{equation}
Replacing fragmented features with the consolidated $Z_{cls}$ reduces the forward passes per denoising step from five to three. This drastically cuts inference memory and latency while strictly preserving world dynamics.

\begin{figure*}[tb!]
    \centering
    \includegraphics[width=\linewidth]{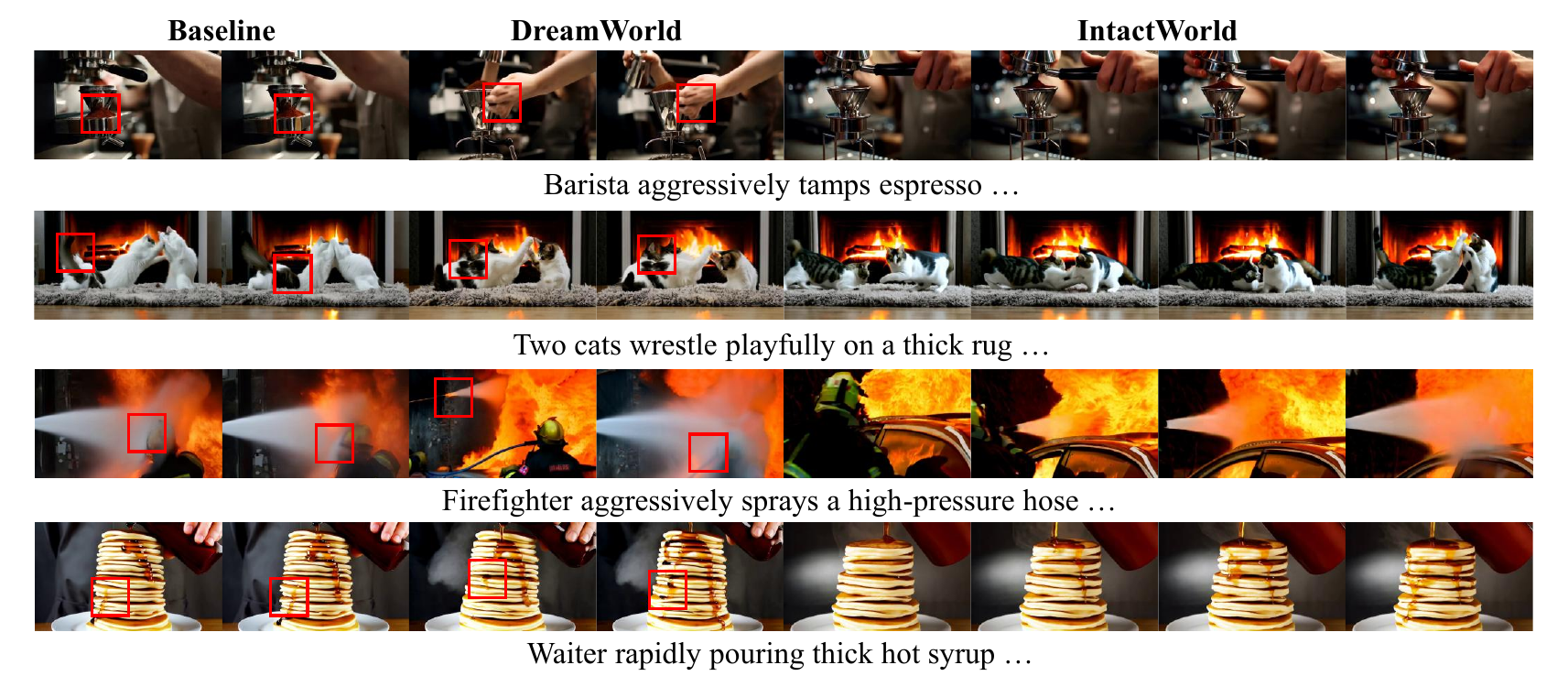}
    \caption{\textbf{Qualitative comparison.} Comparison of IntactWorld with Baseline and DreamWorld where red boxes denote structural errors and artifacts. Notably, IntactWorld achieves superior visual quality with fewer artifacts.}
    \label{fig:Qualitative}
\end{figure*}

\section{Experimental Results}
\subsection{Implementation Details}
% \vspace{-4pt}
\label{subsec:implementation}
\paragraph{Dataset Details.} We conduct our experiments on a 30K subset of the open-source WISA-80K~\cite{wang2025wisa} dataset. All video clips are uniformly sampled to 81 frames at a spatial resolution of $480 \times 832$. Optical flow features are extracted for the entire 30K subset~\cite{chefer2025videojamjointappearancemotionrepresentations}. To support our Full-to-Compact Training Paradigm, we extract uncompressed full DINOv2~\cite{oquab2023dinov2} and VGGT~\cite{wang2025vggtvisualgeometrygrounded} features for 24K videos to train Stage \Rmnum{1}, while extracting only the compact CLS tokens for the remaining 6K videos specifically for Stage \Rmnum{2}.
\paragraph{Training Configuration.} We fine-tune the foundational Wan2.1-T2V-1.3B~\cite{wan2025wanopenadvancedlargescale} model using 8 NVIDIA A100 (80GB) GPUs. The optimization utilizes a learning rate of 1e-5 with a per-device batch size of 2. Following our Full-to-Compact Training Paradigm, Stage \Rmnum{1} is trained for 1,500 steps, after which the learned LoRA weights are saved. Stage \Rmnum{2} is then initialized directly from these Stage \Rmnum{1} LoRA weights and trained for an additional 750 steps with batch size of 1.
\paragraph{Baseline Methods.} We employ the Wan2.1-T2V-1.3B model as our primary baseline, which is directly fine-tuned on the aforementioned 30K dataset using the finetrainers framework. This serves as a standard reference to evaluate the effectiveness of our proposed framework.

\begin{figure*}[tb!]
    \centering
    \begin{minipage}{0.48\linewidth}
        \centering
        \includegraphics[width=\linewidth]{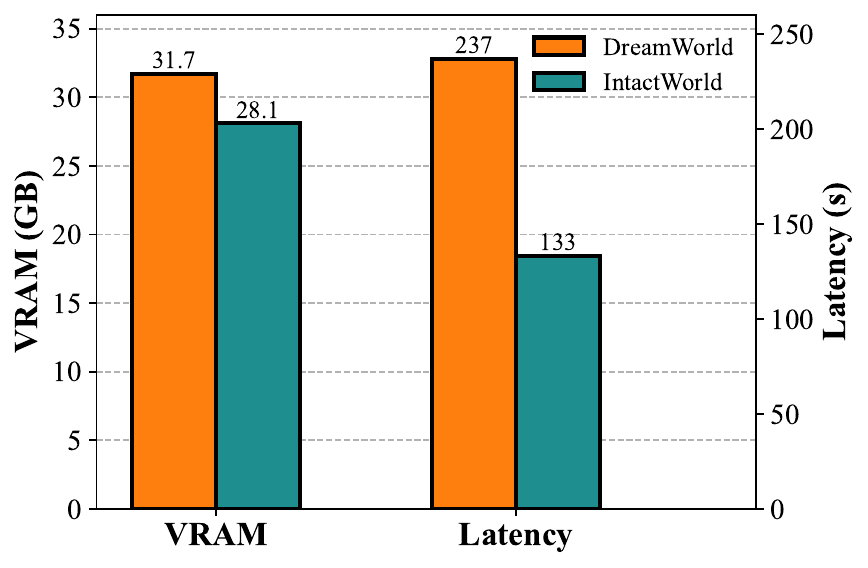}
        \caption{\textbf{Inference Cost.} Comparison of inference latency and VRAM usage between IntactWorld and DreamWorld.}
        \label{fig:inference_cost}
    \end{minipage}
    \hfill
    \begin{minipage}{0.49\linewidth}
        \centering
        \includegraphics[width=\linewidth]{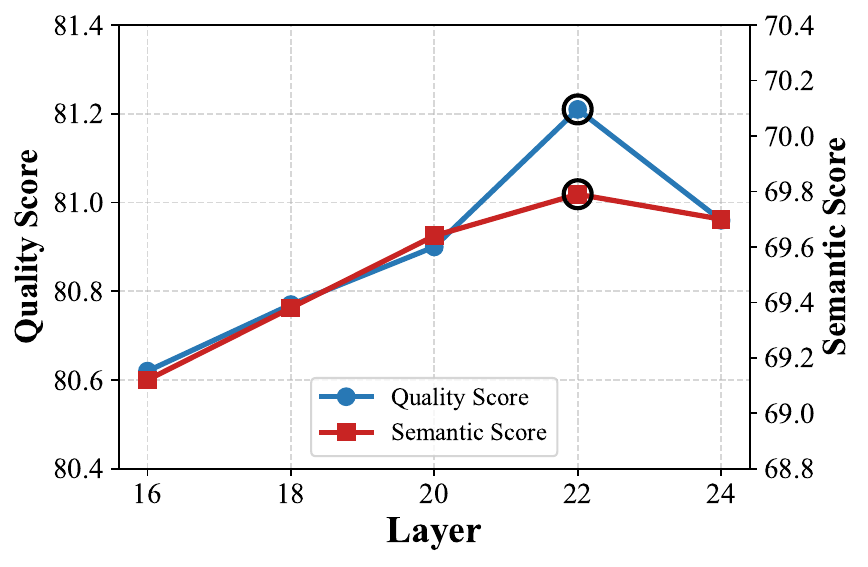}
        \caption{\textbf{Influence of Prediction Depth.} The best generation quality is achieved at layer \textbf{22}.}
        \label{fig:layer_ablation}
    \end{minipage}
\end{figure*}

\subsection{Quantitative Results}
\label{subsec:quantitative}

To evaluate the world modeling effectiveness and overall generation quality of IntactWorld, we conduct quantitative experiments on the authoritative benchmarks VBench \cite{huang2023vbenchcomprehensivebenchmarksuite} and VBench-2.0 \cite{zheng2025vbench20advancingvideogeneration}, as well as VideoPhy \cite{bansal2024videophyevaluatingphysicalcommonsense} specifically for assessing physical commonsense. All text prompts used for video generation are directly sourced from their official repositories.

\paragraph{VBench.}
We evaluate general video synthesis quality using VBench~\cite{huang2023vbenchcomprehensivebenchmarksuite}. As shown in Table~\ref{tab:vbench}, IntactWorld achieves a state-of-the-art overall score of \textbf{81.45}, outperforming the Wan2.1-T2V-1.3B baseline and DreamWorld. Our model particularly excels in the Semantic and Spatial dimensions (e.g., Object Class and Spatial Relationship). This demonstrates that fully utilizing uncompressed features effectively prevents the loss of crucial structural information common in compression-based approaches. 

\paragraph{VBench-2.0.}
Moving beyond surface-level visual aesthetics, VBench-2.0~\cite{zheng2025vbench20advancingvideogeneration} rigorously evaluates the holistic world-modeling capabilities of generative models. On this benchmark, IntactWorld achieves a new state-of-the-art overall score of \textbf{53.64}, significantly surpassing existing baselines (Table~\ref{tab:vbench2}). This demonstrates that by predicting the clean world features $x_0$ and bridging the manifold gap, IntactWorld goes beyond superficial video synthesis; it successfully internalizes comprehensive structural priors to accurately simulate the underlying rules of the real world.

\paragraph{VideoPhy.}
VideoPhy~\cite{bansal2024videophyevaluatingphysicalcommonsense} evaluates physical commonsense across diverse material interactions (solid and fluid scenarios). Table~\ref{tab:videophy} shows that IntactWorld achieves the highest Semantic Adherence (\textbf{55.2}) and Physical Commonsense (\textbf{27.6}) scores, consistently outperforming the baselines. This confirms that our intact feature approach successfully internalizes robust physical priors, allowing the model to simulate genuine material dynamics rather than merely memorizing superficial visual patterns.

\paragraph{Inference Cost.} 
Compared with the multi-source guidance in DreamWorld, our Compact Inner-Guidance significantly optimizes computational efficiency, as shown in Figure \ref{fig:inference_cost}. By eliminating redundant branches, the GPU memory footprint is reduced by 3.6 GB, representing an 11.4\% decrease in memory overhead. Furthermore, the generation time per video is cut from 237 to 133 seconds, achieving a substantial 43.8\% reduction in inference latency. These results demonstrate that IntactWorld provides a more efficient solution for high-fidelity world modeling.

\subsection{Qualitative Results}
\label{Qualitative}

Figure \ref{fig:Qualitative} demonstrates the qualitative performance of IntactWorld compared to the Baseline and DreamWorld in various dynamic settings. The red boxes highlight where prior methods suffer from shape deformation and unnatural artifacts during rapid movement or complex interaction. IntactWorld exhibits robust world-modeling capability by preserving object structure and temporal coherence throughout the sequence. These results indicate that our approach can generate high-fidelity content while alleviating common visual failures observed in prior methods.

\subsection{Ablation Studies}
\label{subsec:ablation}

\paragraph{Effectiveness of Full Feature Training.} 
We evaluate the impact of Stage \Rmnum{1} by bypassing full feature learning and directly training exclusively with CLS tokens. As shown in Table~\ref{tab:ablation}, omitting this foundational stage degrades both video quality and semantic alignment. This confirms that initial full feature training is critical for establishing a robust structural prior before transitioning to the CLS token abstraction.

\paragraph{Effectiveness of CLS Token Abstraction.} 
To verify the necessity of CLS token training, we ablate our method by merging the three raw full features into a 3-branch inference structure without CLS abstraction. As Table~\ref{tab:ablation} illustrates, this variant underperforms across all metrics. This indicates that naively merging high-dimensional raw features causes interference during denoising. In contrast, IntactWorld's refined CLS tokens eliminate inter-feature conflicts, achieving superior intrinsic faithfulness through efficient guidance.

\paragraph{Effectiveness of Predicting $x_0$.} 
We investigate the prediction target by substituting the clean signal $x_0$ with the flow velocity $v$ during intermediate prediction. Although predicting velocity is standard in flow matching, results in Table~\ref{tab:ablation} show a clear performance drop compared to the standard IntactWorld. This suggests that directly supervising $x_0$ provides a more reliable and deterministic signal, which is crucial for preserving structural integrity during feature joint modeling.

\paragraph{Influence of Prediction Depth.} The choice of the layer for intermediate feature prediction is crucial. We evaluate IntactWorld on a 10K dataset across various prediction depths (layers 16, 18, 20, 22, and 24). As depicted in Figure~\ref{fig:layer_ablation}, generation quality consistently improves as prediction moves deeper, peaking at layer 22 before degrading at layer 24. This trend suggests that moderately deep layers strike the best balance between preserving fine-grained structure and capturing semantically meaningful world representations. 

\begin{table}[tb!]
    \centering
    \begin{minipage}[t]{0.5\textwidth}
        \centering
        \renewcommand{\arraystretch}{1.1}
        \begin{tabular}{lccc}
            \toprule
            Method &  Quality & Semantic & Total \\
            \midrule
            w/o full feature  & 83.60 & 71.80 & 81.24 \\
            w/o cls tokens    & 83.64 & 72.02 & 81.31 \\
            pred\_v           & 83.16 & 70.50 & 80.63 \\
            \midrule
            \textbf{IntactWorld}  & \textbf{83.76} & \textbf{72.19} & \textbf{81.45} \\
            \bottomrule
        \end{tabular}
    \end{minipage}
    \caption{\textbf{Ablation Study.} Quantitative evaluation on VBench for full feature training, CLS token abstraction, and the prediction target of $x_0$ versus flow velocity $v$.}
    \label{tab:ablation}
\end{table}
\section{Conclusion}
\label{sec:conlusion}
In this paper, we present \textbf{IntactWorld}, a \textbf{Joint World Modeling Architecture} that effectively internalizes intact world features. To overcome the manifold gap bottleneck in high-dimensional spaces, our method predicts the clean feature $x_0$ at intermediate network layers and computes the optimization loss within the $v$-space. Furthermore, we drastically reduce inference costs by introducing a \textit{Full-to-Compact Training Paradigm} and a \textit{Compact Inner-Guidance mechanism}. By abstracting complex world knowledge into highly condensed CLS tokens, our model achieves highly efficient generation requiring only a single additional feature branch alongside standard classifier-free guidance. Extensive experiments demonstrate that IntactWorld successfully eliminates redundant computational overhead while establishing new records on both VBench and VBench-2.0 benchmarks.

\bibliography{aaai2027}

\end{document}